# DTW K-MEANS CLUSTERING FOR FAULT DETECTION IN PHOTOVOLTAIC MODULES

**Edgar Hernando Sepúlveda Oviedo (1,2), Louise Travé-Massuyès (1,3), Audine Subias (1), Marko Pavlov (2), Corinne Alonso (1)**

1. LAAS-CNRS, Université fédérale de Toulouse, CNRS, INSA, UPS, France
2. Feedgy Solar, France
3. ANITI, Université fédérale de Toulouse, France;

*Núcleo Temático: Energías limpias*

## Introduction

The increase in the use of photovoltaic (PV) energy in the world has shown that the useful life and maintenance of a PV plant directly depend on the ability to quickly detect severe faults on a PV plant. To solve this problem of detection, data based approaches have been proposed in the literature. However, these previous solutions consider only specific behavior of one or few faults. Most of these approaches can be qualified as supervised, requiring an enormous labelling effort (fault types clearly identified in each technology). In addition, most of them are validated in PV cells or one PV module. That is hardly applicable in large-scale PV plants considering their complexity. Alternatively, some unsupervised well-known approaches based on data try to detect anomalies but are not able to identify precisely the type of fault. The most performant of these methods do manage to efficiently group healthy panels and separate them from faulty panels. In that way, this article presents an unsupervised approach called DTW K-means. This approach takes advantages of both the dynamic time warping (DWT) metric [Jun 2011, Tanaka 2016] and the K-means clustering algorithm as a data-driven approach [Niennattrakul 2007, Huang 2016]. The results of this mixed method in a PV string are compared to diagnostic labels established by visual inspection of the panels.

## Methods

The approach proposed in this article is composed of three phases: *i)* Acquisition of the electric current signals; *ii)* Feature Extraction using the DTW metric; and *iii)* Clustering using the K-means algorithm. In the first phase, only the electrical current signal of each panel is captured with a sampling time of one minute. In a second phase, a distance matrix between all the captured current time series is calculated. The DTW metric allows for a comparison of two time series even if they are out of phase. The result of DTW is used as input of the K-means algorithm. The K-means algorithm receives as a parameter the number of clusters *k* to build and, based on this, the centroid of each cluster is calculated and the panels are assigned to the different clusters. In this article, only two clusters are needed to discriminate the PV panels. Figure 1 presents the results obtained with the DTW K-means algorithm.

## Results

Figure 1 presents examples of results obtained with the approach named the DTW K-means on the PV current signals during one day. Figure 1a shows data constituted by 12 PV different current signals.

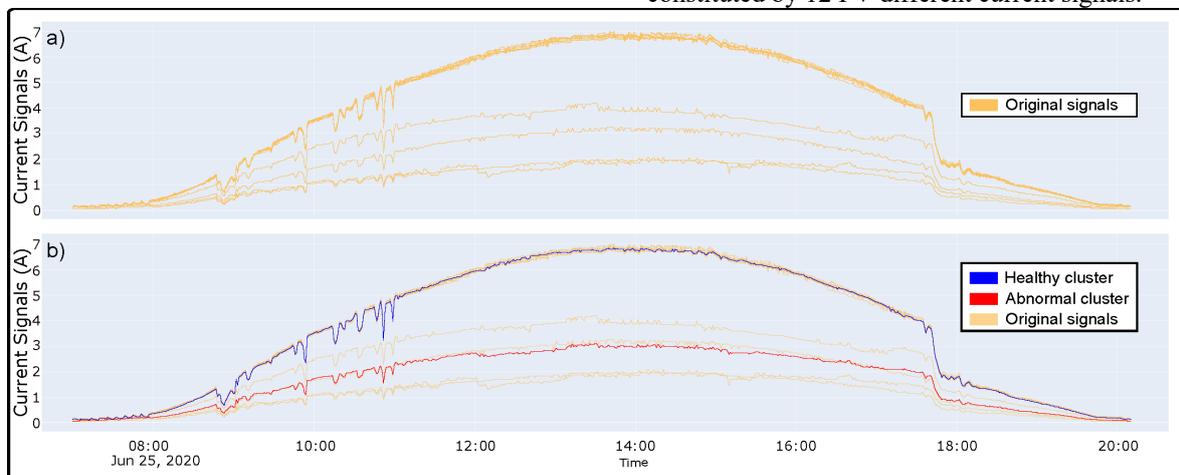

*Figure 1: Clustering results using the DTW K-means. a) Original current signals; b) The two centroids found by the DTW K-means method overlapping on the current signals.*

Figure 1b shows the two centroids found by DTW K-means method overlapping on the current signals. Results have been validated by visual inspection on a real experimental test bench. It was found that all the signals grouped in the healthy cluster (blue color) effectively correspond to 8 healthy panels. For the rest, the 4 panels grouped in the red cluster (abnormal cluster) belong to panels with broken glass.

## Discussion

Figure 1 illustrates that DTW K-means approach that is able to group all healthy panels in one cluster and faulty ones in another cluster correctly. However, after performing the visual inspection, it was found that in some of the healthy panels, there were some snail trail fault types, which are small discolorations. This type of fault is considered as major by PV plant operators because in few months it induces a great degraded electrical performance. The difficulty for the DTW K-means was that at the time of the data processing, this fault did not reduce yet the power production of the panel. For that reason, it would be interesting to explore fine-grained diagnostics that could identify such faults, perhaps using supervised approaches.

Another interesting aspect to highlight about this approach is that it does not require the extraction of multiple characteristic features, which is interesting from the computational cost point of view. In addition, it only requires the current signal from the panels, so the cost in terms of sensors is also reduced. Finally, this approach can be extended to different time windows, the approach was tested in windows of up to 3 minutes, presenting equally consistent results.

## References

B.H. Jun. Fault detection using dynamic time warping (DTW) algorithm and discriminant analysis for swine wastewater treatment. Journal of Hazardous Materials, vol. 185, no. 1, pages 262–268, 2011.

John A. Tsanakas, Long Ha and Claudia Buerhop. Faults and infrared thermographic diagnosis in operating c-Si photovoltaic modules: A review of research and future challenges. Renewable and Sustainable Energy Reviews, vol. 62, pages 695–709, 2016.

X. Huang, Yunming Ye, L. Xiong, Raymond Y. K. Lau, Nan Jiang and Shaokai Wang. Time series k-means: A new k-means type smooth subspace clustering for time series data. Inf. Sci., vol. 367-368, pages 1–13, 2016.

Vit Niennattrakul and Chotirat Ann Ratanamahatana. On Clustering Multimedia Time Series Data Using K-Means and Dynamic Time Warping. In 2007 International Conference on Multimedia and Ubiquitous Engineering (MUE'07), pages 733–738, 2007.